\documentclass{article}
\usepackage{amsfonts}
\usepackage{booktabs}
\usepackage{multirow}
\usepackage{spconf,amsmath,graphicx}
\usepackage{color}

\newcommand{\x}{\mathbf{x}}
\newcommand{\sss}{\mathbf{s}}
\newcommand{\ccc}{\mathbf{c}}
\newcommand{\hh}{\mathbf{h}}
\newcommand{\bfblue}[1]{\textcolor{blue}{\textbf{#1}}}
\newcommand{\parheader}[1]{{\smallskip \bf \noindent #1.}}

\newcommand{\ignore}[1]{}

\title{Streaming Voice Query Recognition using Causal Convolutional Recurrent Neural Networks}
\name{Raphael Tang\sthanks{Work done while interning at Comcast Labs in Washington, D.C.}$^1$ \quad Gefei Yang$^2$ \quad Hong Wei$^*$$^2$ \quad Yajie Mao$^2$ 
\quad Ferhan Ture$^2$ \quad Jimmy Lin$^1$}
\address{$^1$David R. Cheriton School of Computer Science, University of Waterloo\\
$^2$Comcast Applied AI Research Lab, Comcast Corporation\\
  \texttt{\small\{r33tang,jimmylin\}@uwaterloo.ca}, \texttt{\small\{gefei\_yang,hong\_wei,yajie\_mao,ferhan\_ture\}@comcast.com}}
\begin{document}
%
\maketitle
\begin{abstract}
Voice-enabled commercial products are ubiquitous, typically enabled by lightweight on-device keyword spotting (KWS) and full automatic speech recognition (ASR) in the cloud.
ASR systems require significant computational resources in training and for inference, not to mention copious amounts of annotated speech data.
KWS systems, on the other hand, are less resource-intensive but have limited capabilities.
On the Comcast Xfinity X1 entertainment platform, we explore a middle ground between ASR and KWS: 
We introduce a novel, resource-efficient neural network for voice query recognition that is much more accurate than state-of-the-art CNNs for KWS, yet can be easily trained and deployed with limited resources.
On an evaluation dataset representing the top 200 voice queries, we achieve a low false alarm rate of 1\% 
and a query error rate of 6\%. Our model performs inference 8.24$\times$ faster than the current ASR system.

\end{abstract}
\section{Introduction}
\label{sec:intro}

Most voice-enabled intelligent agents, such as Apple's Siri and the Amazon Echo, are powered by a combination of two technologies:\ lightweight keyword spotting (KWS) to detect a few pre-defined phrases within streaming audio (e.g., ``Hey Siri'') and full automatic speech recognition (ASR) to transcribe complete user utterances.
In this work, we explore a middle ground:\ techniques for voice query recognition capable of handling a couple of hundred commands.

Why is this an interesting point in the design space?
On the one hand, this task is much more challenging than the (at most) a couple of dozen keywords handled by state-of-the-art KWS systems~\cite{tang2018deep, arik2017convolutional}.
Their highly constrained vocabulary limits application to wake-word and simple command recognition.
Furthermore, their use is constrained to detecting whether some audio contains a phrase, not exact transcriptions needed for voice query recognition.
For example, if ``YouTube'' were the keyword, KWS systems would make no distinction between the phrases ``quit YouTube'' and ``open YouTube''---this is obviously not sufficient since they correspond to different commands.
On the other hand, our formulation of voice query recognition was specifically designed to be far more lightweight than full ASR models, typically recurrent neural networks that comprise tens of millions of parameters, take weeks to train and fine tune, and require enormous investment in gathering training data.
Thus, full ASR typically incurs high computational costs during inference time and have large memory footprints~\cite{chiu2017state}.

The context of our work is the Comcast Xfinity X1 entertainment platform, which provides a ``voice remote'' that accepts spoken queries from users.
A user, for example, might initiate a voice query with a button push on the remote and then say ``CNN'' as an alternative to remembering the exact channel number or flipping through channel guides.
Voice queries are a powerful feature, since modern entertainment packages typically have hundreds of channels and remote controls have become too complicated for many users to operate.
On average, X1 accepts tens of millions of voice queries per day, totaling 1.7 terabytes of audio, equal to 15,000 spoken hours.

A middle ground between KWS and full ASR is particularly interesting in our application because of the Zipfian distribution of users' queries.
The 200 most popular queries cover a significant portion of monthly voice traffic and accounts for millions of queries per day.
The key contribution of this work is a novel, resource-efficient architecture for streaming voice query recognition on the Comcast X1.
We show that existing KWS models are insufficient for this task, and that our models answer queries more than eight times faster than the current full ASR system, with a low false alarm rate (FAR)\ of 1.0\% and query error rate (QER) of 6.0\%.

\begin{figure*}[!ht]
    \centering
    \includegraphics[scale=0.7]{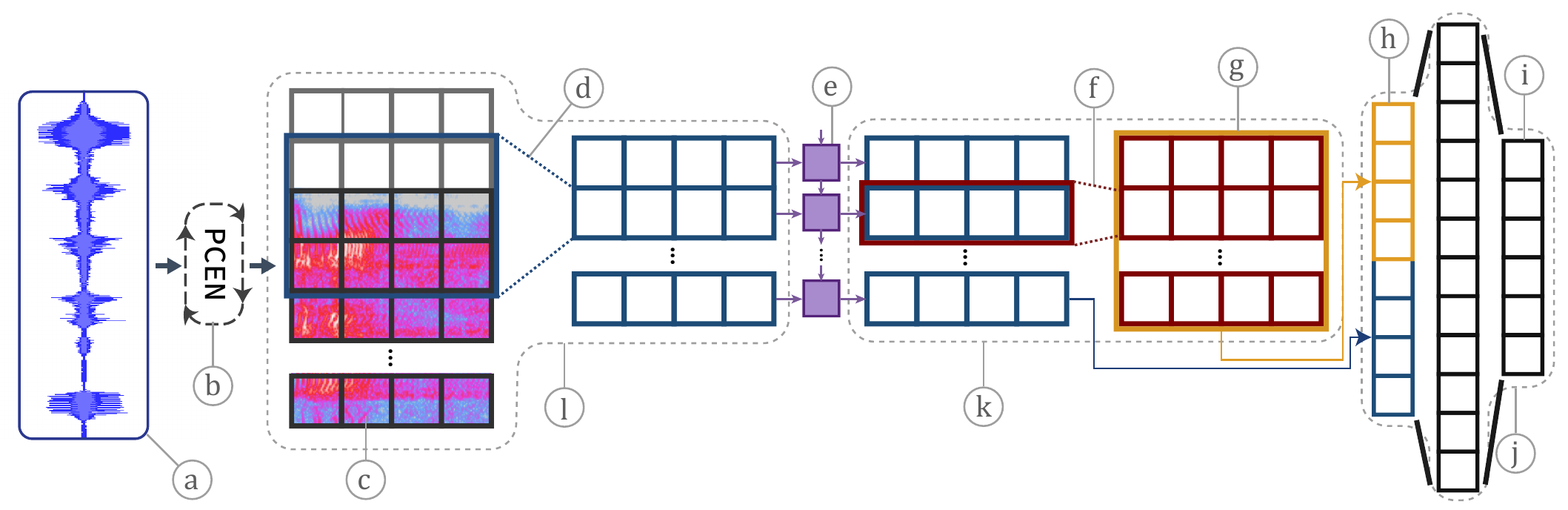}\vspace{-3mm}
    \caption{Illustration of our architecture. The labels are as follows: \textbf{(A)} raw audio waveform \textbf{(B)} 
        streaming Mel--PCEN filterbank \textbf{(C)} PCEN features
        \textbf{(D)} causal convolution \textbf{(E)}~GRU layer \textbf{(F)} feature extraction convolution \textbf{(G)}
        max-pool across time \textbf{(H)} output concatenation \textbf{(I)} 201-class output \textbf{(J)} DNN classifier
        \textbf{(K)} long-term context modeling \textbf{(L)} short-term context modeling.}
    \label{fig:model_arch}
\end{figure*}
\section{Related Work}
The typical approach to voice query recognition is to develop a full automatic
speech recognition (ASR) system~\cite{las}. Open-source toolkits like 
Kaldi~\cite{Povey:192584} provide ASR models to researchers; however, 
state-of-the-art commercial systems frequently require thousands of hours of
training data~\cite{xiong2017microsoft} and dozens of gigabytes for the 
combined acoustic and language models~\cite{chiu2017state}. Furthermore, 
we argue that these systems are excessive for usage scenarios characterized by Zipf's
Law, such as those often encountered in voice query recognition:\ for example, on 
the X1, the top 200 queries cover a significant, disproportionate amount of our entire
voice traffic. Thus, to reduce computational requirements associated with training
and running a full ASR system, we propose to develop a lightweight model for handling the 
top-K queries only.

While our task is related to keyword spotting, KWS systems only
strictly detect the mere occurrence of a phrase within audio, not the exact transcription,
as in our task. Neural networks with both convolutional and recurrent components have 
been successfully used in keyword spotting~\cite{arik2017convolutional, gclstm}; others use only
convolutional neural networks~(CNNs)~\cite{sainath2015convolutional, tang2018deep} and 
popular image classification models~\cite{mcmahan2018}.

\section{Task and Model}

Our precise task is to classify an audio clip as one of $N + 1$ classes, 
with $N$ labels denoting $N$ different voice queries and a single unknown 
label representing everything else. To improve responsiveness and hence the user experience, 
we impose the constraint that model inference executes in an on-line, streaming manner, 
defined as predictions that occur every 100 milliseconds and in constant time and space, 
with respect to the total audio input length. This enables software applications to display on-the-fly
transcriptions of real-time speech, which is important for user satisfaction:\ we immediately begin processing 
speech input when the user depresses the trigger button on the X1 voice remote.

\subsection{Input preprocessing}
First, we apply dataset augmentation to reduce generalization error in speech recognition models~\cite{ko2015audio}. In our work,
we randomly apply noise, band-pass filtering, and pitch shifting to each audio sample. Specifically, we add 
a mixture of Gaussian and salt-and-pepper noise---the latter is specifically chosen due to the voice remote
microphone introducing such artifacts, since we notice ``clicks'' while listening to audio samples. For band-pass 
filtering, we suppress by a factor of 0.5 the frequencies outside a range with random endpoints $[a, b]$, where $a$ and $b$ 
roughly correspond to frequencies drawn uniformly from $[0, 1.7]$ kHz and $[1.8, 3.3]$ kHz, respectively. For pitch shifting, 
we apply a random shift of $\pm$33 Hz. The augmentation procedure was verified by ear to be reasonable.

We then preprocess the dataset from raw audio waveform to forty-dimensional 
per-channel energy normalized (PCEN)~\cite{wang2017trainable} frames, with a window size of 30 milliseconds
and a frame shift of 10 milliseconds. PCEN provides robustness to per-channel energy differences between 
near-field and far-field speech applications, where it is used to achieve the state of the art in keyword 
spotting~\cite{arik2017convolutional, wang2017trainable}. Conveniently, it handles
streaming audio; in our application, the user's audio is streamed in real-time to our platform. 
As is standard in speech recognition applications, all audio is recorded in 16kHz, 
16-bit mono-channel PCM format.

\subsection{Model Architecture}

We draw inspiration from convolutional recurrent neural 
networks (ConvRNN) for text modeling~\cite{convrnn}, where it 
has achieved state of the art in sentence classification. 
However, the model cannot be applied as-is to our task, since the bi-directional 
components violate our streaming constraint, and it was originally designed for no 
more than five output labels. Thus, we begin with this model as a template only.

We illustrate our architecture in Figure \ref{fig:model_arch}, where the model can be best described as having three sequential components: first, it
uses causal convolutions to model short-term speech context. Next, it feeds the short-term context
into a gated recurrent unit (GRU)~\cite{cho2014learning} layer and pools across time to model long-term context. Finally, it feeds
the long-term context into a deep neural network (DNN) classifier for our $N + 1$ voice query labels.

\parheader{Short-term context modeling}
Given 40-dimensional PCEN inputs $\x_1, \dots, \x_t$, we first stack the frames to form 
a 2D input $\x_{1:t} \in \mathbb{R}^{t\times 40}$; see Figure \ref{fig:model_arch}, label C, where
the \textit{x}-axis represents 40-dimensional features and the \textit{y}-axis time. Then, to model short-term context, we use a 2D causal 
convolution layer (Figure \ref{fig:model_arch}, label D) to extract feature vectors $\sss_1, \dots, \sss_t$ for 
$\sss_{1:t} = W \cdot \x + b$, where $W \in \mathbb{R}^{c \times (m \times n)}$ is the convolution
weight, $\x_{-m+2:0}$ is silence padding in the beginning, $\cdot$ denotes valid convolution, and $\sss_i$ is
a context vector in $\mathbb{R}^{c \times f}$. Finally, we pass the outputs into a rectified linear (ReLU) activation 
and then a batch normalization layer, as is standard in image classification. Since causal convolutions use a 
fixed number of past and current inputs only, the streaming constraint is necessarily maintained. 

\parheader{Long-term context modeling}
To model long-term context, we first flatten the short-term context vector per time step from 
$\sss_i \in \mathbb{R}^{c \times f}$ to $\mathbb{R}^{cf}$. Then, we feed them into a 
single uni-directional GRU layer (examine Figure \ref{fig:model_arch}, label E) consisting of $k$ hidden units, yielding hidden outputs
$\hh_1, \dots, \hh_t, \hh_i \in \mathbb{R}^{k}$. Following text modeling work~\cite{convrnn}, we
then use a 1D convolution filter $W \in \mathbb{R}^{d \times k}$ with ReLU activation to extract 
features from the hidden outputs, where $d$ is the number of output channels. We max-pool these features 
across time (see Figure \ref{fig:model_arch}, label G) to obtain a fixed-length context $\ccc_{\text{max}} \in \mathbb{R}^{d}$. Finally, we concatenate
$\ccc_{\text{max}}$ and $\hh_t$ for the final context vector, $\ccc \in \mathbb{R}^{k + d}$, as shown in
Figure \ref{fig:model_arch}, label H.

Clearly, these operations maintain the streaming constraint, since uni-directional GRUs and max-pooling 
across time require the storage of only the last hidden and maximum states, respectively. We also 
experimentally find that the max-pooling operation helps to propagate across time the strongest 
activations, which may be ``forgotten'' if only the last hidden output from the GRU were used as 
the context. 

\parheader{DNN classifier}
Finally, we feed the context vector $\ccc$ into a small DNN with one hidden layer with ReLU activation, and a 
softmax output across the $N + 1$ voice query labels. For inference on streaming audio, we merely execute the 
DNN on the final context vector at a desired interval, such as every 100 milliseconds; in our models, we 
choose the number of hidden units $r$ so that the classifier is sufficiently lightweight.

\section{Evaluation}

On our specific task, we choose $N = 200$ representing the top 200 queries on the Xfinity X1 platform,
altogether covering a significant portion of all voice traffic---this subset corresponds to
hundreds of millions of queries to the system per month. For each positive class, we 
collected 1,500 examples consisting of anonymized real data. For the negative class, 
we collected a larger set of 670K examples not containing any of the positive keywords. Thus, our dataset 
contains a total of 970K examples. For the training set, we used the first 80\% of each class; for the 
validation and test sets, we used the next two 10\% partitions. Each example was extremely short---only
2.1 seconds on average. All of the transcriptions were created by a state-of-the-art commercial 
ASR system with 5.8$\pm 1.6$\% (95\% confidence interval) word-error rate (WER) on our dataset; this choice 
is reasonable because the WER of human annotations is similar~\cite{stolcke2017comparing}, and our deployment approach
is to short-circuit and replace the current third-party ASR system where possible.

\subsection{Training and Hyperparameters}
\begin{table}[t]
    \centering
    \setlength{\tabcolsep}{3.5pt}
    \begin{tabular}{l l c c c}
        \toprule[1pt]
        \# & Type & \# Par. & \# Mult. & Hyperparameters\\
        \midrule\addlinespace[-0.05mm] 
        \multicolumn{5}{c}{\small \bf Short-term context modeling \vspace{-0.5mm}}\\
        \midrule
        1\hspace{1mm} & C. Conv & 15K & 4.5M & $c, m, n=250, 3, 20$\\
        2             & BN & 500 & 150K & --\\
        \midrule\addlinespace[-0.05mm] 
        \multicolumn{5}{c}{\small \bf Long-term context modeling \vspace{-0.5mm}}\\
        \midrule
        3             & GRU & 3.38M & 337M & $k=750$\\
        4             & Conv & 263K & 26.2M & $d=350$\\
        \midrule\addlinespace[-0.05mm] 
        \multicolumn{5}{c}{\small \bf DNN classifier (100ms interval) \vspace{-0.5mm}}\\
        \midrule
        5             & DNN & 845K & 8.4M & $r=768$\\
        6             & Softmax & 154K & 1.5M & $N + 1=201$\\
        \midrule[1pt]
        \multicolumn{2}{l}{\bf Total:} & 4.66M & 378M & --\\
        \bottomrule[1pt]
    \end{tabular}
    \caption{Model footprint and hyperparameters. ``\# Mult.'' denotes the number of multiplies for one second of audio.}
    \label{table:hyperparams}
\end{table}

For the causal convolution layer, we choose $c = 250$ output channels, $m = 3$ width in time, and 
$n = 20$ length in frequency. We then stride the entire filter across time and frequency by one and ten
steps, respectively. This configuration yields a receptive field of 50 milliseconds across $f = 3$ 
different frequency bands, which roughly correspond to highs, mids, and lows. For long-term context modeling, 
we choose $k = 750$ hidden dimensions and $d = 350$ convolution filters. Finally, we choose the hidden layer 
of the classifier to have 768 units. Table \ref{table:hyperparams} summarizes the footprint and hyperparameters of our 
architecture; we name this model \texttt{crnn-750m}, with the first ``\texttt{c}'' representing the causal 
convolution layer and the trailing ``\texttt{m}'' max pooling.

During training, we feed only the final context vector of the entire audio sample into the 
DNN classifier. For each sample, we obtain a single softmax output across the 201 targets for the cross 
entropy loss. The model is then trained using stochastic gradient descent with a momentum of 0.9, batch size 
of 48, $L_2$ weight decay of $10^{-4}$, and an initial learning rate of $10^{-2}$. At epochs 9 
and 13, the learning rate decreases to $10^{-3}$ and $10^{-4}$, respectively, before training finishes 
for a total of 16 epochs.

\parheader{Model Variants}
As a baseline, we adapt the previous state-of-the-art KWS model \texttt{res8}~\cite{tang2018deep} to our task by increasing the number of outputs
in the final softmax to 201 classes. This model requires fixed-length audio, so we pad and trim audio input to a length that is sufficient
to cover most of the audio in our dataset. We choose this length to be eight seconds, since 99.9\% of queries are shorter.

To examine the effect of the causal convolution layer, we train a model without it, feeding the PCEN 
inputs directly to the GRU layer. We also examine the contribution of max-pooling across time by removing it: we 
name these variants \texttt{rnn-750m} and \texttt{crnn-750}.

\subsection{Results and Discussion}

\begin{table}[t]
    \centering
    \setlength{\tabcolsep}{2pt}
    \small
    \begin{tabular}{l l c c c c c c c}
        \toprule[1pt]
        \multirow{2}{*}{\raisebox{-3\heavyrulewidth}{\bf \#}} &
        \multirow{2}{*}{\raisebox{-3\heavyrulewidth}{\bf Model}} & 
        \multicolumn{2}{c}{\bf Val.} &
        \multicolumn{2}{c}{\bf Test} &
        \multicolumn{2}{c}{\bf Footprint}\\
        \cmidrule(lr){3-4} 
        \cmidrule(lr){5-6}
        \cmidrule(lr){7-8}
        & & FAR & QER & FAR & QER & \# Par. & \# Mult.\\
        \midrule
        1\hspace{1mm} & \texttt{res8} & 1.0\% & 29.4\% & 0.9\% & 29.2\% & 110K & 240M\\
        \midrule
        2 & \texttt{crnn-750m} & 1.0\% & \bfblue{6.0\%} & 1.0\% & \bfblue{6.0\%} & 4.66M & 378M\\
        3 & \texttt{crnn-750} & 1.0\% & 6.4\% & 1.0\% & 6.5\% & 4.39M & 354M\\
        4 & \texttt{rnn-750m} & 1.0\% & 6.4\% & 1.0\% & 6.3\% & 3.04M & 267M\\
        \bottomrule[1pt]
    \end{tabular}
    \caption{Comparison of model results. ``\# Mult.'' denotes number of multiplies per second of audio. Note that for \texttt{res8}, we report
    the number on eight seconds of audio, since fixed-length input is expected. Best results are bolded.}
    \label{table:results}
\end{table}

The model runs quickly on a commodity GPU machine with one Nvidia GTX 1080: to classify one second of streaming audio, our model takes 68 milliseconds. 
Clearly, the model is also much more lightweight than a full ASR system, occupying only 
19 MB of disk space for the weights and 5 KB of RAM for the persistent state per audio stream.
The state consists of the two previous PCEN frames for the causal convolution layer (320 bytes; all zeros
for the first two padding frames), the GRU hidden state (3 KB), and the last maximum state for max-pooling across time (1.4 KB).

In our system, we define a false alarm (FA) as a negative misclassification. In other words, a model prediction
is counted as an FA if it is misclassified \textit{and} the prediction is one of the known, 200 queries. This is reasonable,
since we fall back to the third-party ASR system if the voice query is classified as unknown. We also define a query error (QE) as any
misclassified example; then, false alarm rate (FAR) and query error rate (QER) correspond to the number of FAs and QEs, respectively, 
divided by the number of examples. Thus, the overall query accuracy rate is $1 - QER$.

\begin{figure}[t]
    \vspace{-4mm}
    \includegraphics[scale=0.54]{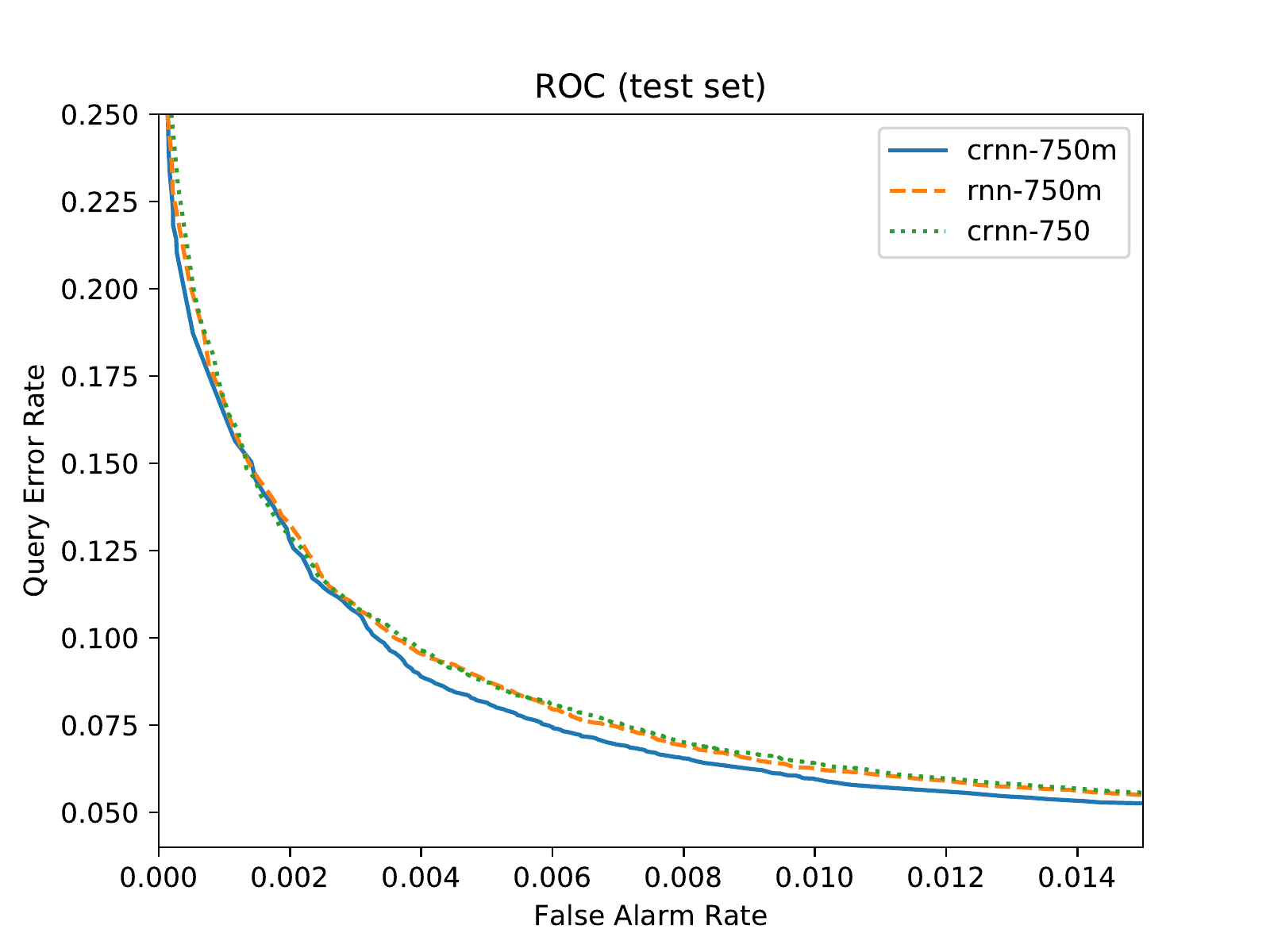}
    \caption{ROC curves for our models.}
    \label{fig:roc}
\end{figure}

Initially, the best model, \texttt{crnn-750m}, attains an FAR and QER of $2.3\%$ and $5.0\%$, respectively. 
This FAR is higher than our production target of $1\%$; thus, we further threshold the predictions 
to adjust the specificity of the model. Used also in our previous work~\cite{tang2018deep}, a simple approach
is to classify as unknown all predictions whose probability outputs are below some global threshold $\alpha$. That is,
if the probability of a prediction falls below some threshold $\alpha$, it is classified as unknown.
In Table \ref{table:results}, we report the results corresponding to our target FAR of $1\%$, with the
$\alpha$ determined from the validation set. To draw ROC curves (see Figure \ref{fig:roc}) on the test set, we sweep $\alpha$ from 0 to 0.9999, where
QER is analogous to false reject rate (FRR) in the classic keyword spotting literature.
We omit \texttt{res8} due to it having a QER of 29\%, which is unusable in practice.

After thresholding, our best model with max pooling and causal convolutions (\texttt{crnn-750m}) achieves an FAR of 1\% and QER of 6\% on both 
the validation and test sets, as shown in Table \ref{table:results}, row 2. Max-pooling across time is effective, resulting in a QER improvement of 0.5\% over
the ablated model (\texttt{crnn-750}; see row 3). The causal convolution layer is effective as well, though slightly less than
max-pooling is; for the same QER (6.4\%) on the validation set, the model without the causal
convolution layer, \texttt{rnn-750m}, uses 87M fewer multiplies per second than \texttt{crnn-750} does (presented in row 4), due to the 
large decrease in the number of parameters for the GRU, which uses an input of size 40 in \texttt{rnn-750m}, compared to 750 in \texttt{crnn-750}.
We have similar findings for the ROC curves (see Figure \ref{fig:roc}), where \texttt{crnn-750m} outperforms \texttt{crnn-750} and \texttt{rnn-750m},
and the ablated models yield similar curves. All of these models greatly outperform \texttt{res8}, which was
originally designed for keyword spotting.

\section{Conclusion and Future Work}

We describe a novel resource-efficient model for the task of voice query recognition on streaming audio,
achieving an FAR and QER of $1\%$ and $6\%$, respectively, while performing more than $8\times$ faster
than the current third-party ASR system. One potential extension to this paper is to explore the application of neural network
compression techniques, such as intrinsic sparse structures~\cite{wen2018learning} and binary quantization~\cite{courbariaux2016binarized}, 
which could further decrease the footprint of our model.

\vfill\pagebreak




\end{document}